\documentclass[10pt]{article} 
\usepackage{amsfonts}
\usepackage{empheq}
\usepackage{graphicx}
\usepackage{subcaption}
\usepackage{url}
\usepackage{amsthm}
\usepackage{amsmath,amssymb}
\usepackage{soul}
\usepackage{wrapfig}
\usepackage{amsmath}
 \usepackage{booktabs} 
\DeclareMathOperator*{\argmax}{argmax}

\usepackage[colorlinks=true,allcolors=perfblue]{hyperref} 

\usepackage[preprint]{rlj} 

\usepackage{amssymb}            
\usepackage{mathtools}          
\usepackage{mathrsfs}           
\usepackage{graphicx}           
\usepackage{subcaption}         
\usepackage[space]{grffile}     
\usepackage{url}                
\usepackage{lipsum}             

\usepackage{amsthm}
\usepackage{thmtools}
\theoremstyle{plain}
\newtheorem{theorem}{Theorem}[section]

\newtheorem{lemma}[theorem]{Lemma}

\theoremstyle{definition}

\theoremstyle{remark}

\declaretheoremstyle[
headfont=\normalfont\itshape,
qed=\qedsymbol,
]{mypf}
\declaretheorem[numbered=no, name=Proof, style=mypf]{pf}

 \usepackage{bbm}

 \usepackage[customcolors]{hf-tikz}
\definecolor{expert}{HTML}{008000}
\definecolor{error}{HTML}{f96565}
\definecolor{learner}{HTML}{F79646}
\definecolor{perfblue}{RGB}{64, 114, 175}

\title{Your Learned Constraint is Secretly a Backward Reachable Tube}

\setrunningtitle{Your Learned Constraint is Secretly a Backward Reachable Tube}

\author{Mohamad Qadri$^{1}$, 
  Gokul Swamy$^{1}$, 
  Jonathan Francis$^{1,2}$,
  Michael Kaess$^{1}$,
  Andrea Bajcsy$^{1}$\\
}

\emails{\{mqadri, gswamy, jmf1, kaess, abajcsy\}@cs.cmu.edu}

\affiliations{
$^{1}$\textbf{Carnegie Mellon University, Robotics Institute}\\
$^{2}$\textbf{Bosch Center for Artificial Intelligence}\\
}

\contribution{
        This paper establishes a connection between Inverse Constraint Learning and Hamilton-Jacobi (HJ) Reachability from safe control theory, providing a new theoretical perspective on learning constraints from demonstrations.
    }
    {
        None
    }

\contribution{
    We prove theoretically and verify experimentally that the mathematical set encoded by the learned constraint is a dynamics-dependent Backward Reachable Tube (BRT) and not the dynamics independent Failure Set.
}
{
    Prior works implicitly assume that the constraint learned via ICL is dynamics independent. In this paper we show that the constraint will actually depend on the dynamics of the expert demonstrators.
}

\contribution{
 We discuss the implication of this observation in terms of the sample-efficiency of policy search and transferability of the learned constraint.
}
{
None
}


\summary{

Inverse Constraint Learning (ICL) is the problem of inferring constraints from safe (i.e., constraint-satisfying) demonstrations. The hope is that these inferred constraints can then be used downstream to search for safe policies for new tasks and, potentially, under different dynamics. Our paper explores the question of what mathematical entity ICL recovers. Somewhat surprisingly, we show that both in theory and in practice, ICL recovers the set of states where failure is \textit{inevitable}, rather than the set of states where failure has \textit{already} happened. In the language of safe control, this means we recover a \textit{backwards reachable tube (BRT)} rather than a \textit{failure set}. In contrast to the failure set, the BRT depends on the dynamics of the data collection system. We discuss the implications of the dynamics-conditionedness of the recovered constraint on both the sample-efficiency of policy search and the transferability of learned constraints. Our code is available in the following  \href{https://github.com/CMU-IntentLab/ICL-BRT}{repository}.
}

\begin{document}

\maketitle  

\begin{abstract}
Inverse Constraint Learning (ICL) is the problem of inferring constraints from safe (i.e., constraint-satisfying) demonstrations. The hope is that these inferred constraints can then be used downstream to search for safe policies for new tasks and, potentially, under different dynamics. Our paper explores the question of what mathematical entity ICL recovers. Somewhat surprisingly, we show that both in theory and in practice, ICL recovers the set of states where failure is \textit{inevitable}, rather than the set of states where failure has \textit{already} happened. In the language of safe control, this means we recover a \textit{backwards reachable tube (BRT)} rather than a \textit{failure set}. In contrast to the failure set, the BRT depends on the dynamics of the data collection system. We discuss the implications of the dynamics-conditionedness of the recovered constraint on both the sample-efficiency of policy search and the transferability of learned constraints. Our code is available in the following  \href{https://github.com/CMU-IntentLab/ICL-BRT}{repository}. 
\end{abstract}

\section{Introduction}
Constraints are fundamental for safe robot decision-making \citep{stooke2020responsive, qadri2022incopt, howell2022calipso}. However, manually specifying safety constraints can be challenging for complex problems, paralleling the \textit{reward design} problem in reinforcement learning \citep{hadfield2017inverse}. For example, consider the example of an off-road vehicle that needs to traverse unknown terrains. Successful completion of this task requires satisfying constraints such as ``avoid terrains that, when traversed, will cause the vehicle to flip over'' which can be difficult to specify precisely via a hand-designed function. Hence, there has been a growing interest in applying techniques analogous to Inverse Reinforcement Learning (IRL) --- where the goal is to learn hard-to-specify reward functions --- to learning constraints \citep{liu2024comprehensive}. This is called Inverse Constraint Learning (ICL): given safe expert robot trajectories and a nominal reward function, we aim to extract the implicit constraints that the expert demonstrator is satisfying. Intuitively, these constraints forbid highly rewarding behavior that the expert nevertheless chose not to take \citep{NEURIPS2023_124dde49}. However, as we now explore, the question of what object we're actually inferring in ICL has a nuanced answer that has several implications for downstream usage of the inferred constraint.

\begin{figure}[h!]
    \centering
`       \includegraphics[width=0.9\linewidth]{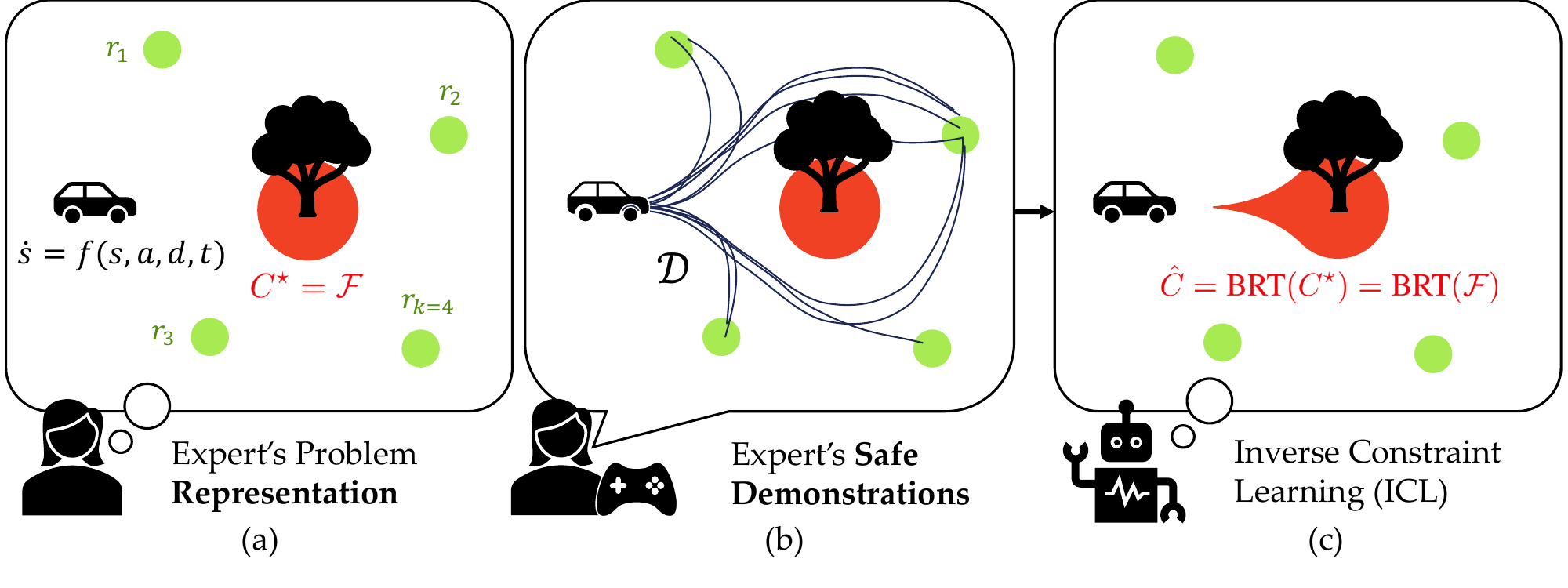}
    \caption{In this work we show theoretically and empirically that inverse constraint learning (ICL) recovers a backwards reachable tube rather than the true failure set as commonly assumed in the literature. (a) ICL models the expert demonstrator as optimizing reward functions (potentially for different tasks) while satisfying a shared true constraint $c^\star$ (e.g. don't hit a tree) with an associated unsafe set $C^\star$. (b) ICL takes as input expert demonstrations and the reward functions $r_1 \hdots r_K$ and aims to recover the shared true constraint $c^\star$. (c) ICL infers a constraint $\hat
    c$ and it's associated unsafe set $\hat{C }$, from the demonstrations. However, we show that $\hat{c}$ encodes a different object than the true failure set. In particular, $\hat{c}$ encodes the the \textit{backward reachable tube} of the true failure set under system dynamics $f(x,a,d,t)$: the set of states from which violating $c^\star$ is inevitable (e.g. positions / velocities for which we can't avoid crashing). \vspace{-0.5cm}}
    \label{fig:simple-example}
\end{figure}

Consider Fig. \ref{fig:simple-example}a , in which an expert (e.g., a human driver) drives a car through a forest from a starting position to an end goal, without colliding with any trees. Assume that the expert has an internal representation of the true constraint, $c^\star$, which they use during their planning process to generate demonstrations (Fig. \ref{fig:simple-example}b).  Here, $c^\star$ encodes the location of the trees or, equivalently, the \textit{failure set}: the set of states which encode having \textit{already} failed the task. Given expert demonstrations that satisfy $c^\star$, we can run an ICL algorithm to obtain an inferred constraint, $\hat{c}$. Our key question is whether the learner actually recovers the constraint the expert used (i.e. is $\hat{c} = c^\star$?). In other words, does $\hat{c}$  encode the true failure set (e.g., where the trees are)?  As we prove below, the answer to this question is, surprisingly enough, often ``\textit{no}." 

This motivates the key question of our work:
\begin{quote}
    \centering
    \textit{When learning constraints from demonstrations, \\ what \textbf{ mathematical entity} are we actually learning?}
\end{quote}


We show theoretically and empirically that, rather than inferring the set of states where the robot has \textit{already} failed at the task, ICL instead infers where, under the expert's dynamics, failure is \textit{inevitable}. For example, rather than inferring the location of the tree, ICL would infer the larger set of states for which the expert will find that avoiding the tree is impossible (illustrated in Fig. \ref{fig:simple-example}c). More formally, we prove that ICL is actually approximating a dynamics-conditioned \textit{backward reachable tube} (BRT), rather than the the dynamics-independent failure set.
The observation that we are recovering \textit{dynamics-conditioned} BRTs rather than failure sets has two important implications. 
On one hand, it means that we can add ICL algorithms to the set of computational tools available to us for computing BRTs, given a dataset of safe demonstrations. 
On the other hand, it means that one cannot hope to easily transfer the constraints learned via ICL between different dynamics naively.

We begin by exploring the relationship between ICL and BRTs before discussing implications.

\section{Problem Setup}
\vspace{-0.2cm}
\noindent \textbf{Dynamical System Model.} 
We consider continuous-time dynamical systems described by the ordinary differential equation \(\dot{s} = f(s, a, d, t)\), where \(t\) is time, \(s \in \mathcal{S}\) is the state, \(a \in \mathcal{A}\) is the control input, and \(d \in \mathcal{D}\) is the disturbance that accounts for unmodeled dynamics (e.g., wind or friction). We note that the 4-tuple $(\mathcal S,\mathcal A,\mathcal D,f)$ induces a Markov decision process (MDP) in continuous time: for any state–action–disturbance triple $(s,a,d)$, the next-state distribution is deterministic and given by the unique solution of the ODE at an infinitesimal time later.  Hence, our setup can be interpreted either in the standard control sense or in the language of MDPs that is familiar to the RL community.

\noindent \textbf{Environment and Task Definition.} 
A task \(k\) is defined as a specific objective that our robot needs to complete. For example, in Fig.~\ref{fig:simple-example}, a mobile robot might be tasked with reaching a specific target pose from a starting position while avoiding environmental obstacles. 
In this work, we assume this task objective to be implicitly defined using a reward function \(r_k: \mathcal{S}\times \mathcal{A}  \rightarrow \mathbb{R}\). Let \(K\) be a set of tasks \(\{k\}\) with a shared implicit constraint $c^\star$ which can be a function of the state and action or of the state only---in other words,
$c^\star: \mathcal{S} \times \mathcal{A} \rightarrow \mathbb{R} $ or $c^\star: \mathcal{S} \rightarrow \mathbb{R}$. 
While the task-specific reward \(r_k\) assigns a high reward when the robot successfully completes the objective, the constraint \(c^\star\) assigns a high cost to state-action pairs (or states) that violate the true shared constraints. In other words,
$c^\star = \infty$ if a state-action pair (or state) is unsafe and $c^\star=-\infty$ otherwise. For example, in Fig. \ref{fig:simple-example}a, the set $C^\star = \{ s \in \mathcal{S} \; \vert \; 1[c^\star(s)=\infty] \}$ represents the true location of the obstacle (i.e., the tree). Furthermore, let's define \(\hat{c}: \mathcal{S} \times \mathcal{A} \rightarrow \mathbb{R}\) or \(\hat{c}: \mathcal{S} \rightarrow \mathbb{R}\) as the constraint learned through ICL. Similarly, \(\hat{c} = \infty\) if a state-action pair (or state) is deemed unsafe by the algorithm and \(\hat{c} = - \infty\) otherwise. For example, in Fig. \ref{fig:simple-example}c, \(\hat{C} = \{ s \in \mathcal{S} \; \vert \; 1[\hat{c}(s)=\infty] \}\) represents the inferred set of unsafe states calculated by ICL.

\noindent \textbf{Safe Demonstration Data from an Expert.} 
In the ICL setting, an expert provides our algorithm with safe demonstrations from \(K\) different tasks, each satisfying a shared constraint. Take, for instance, a mobile robot operating in a single environment as shown  Fig. \ref{fig:simple-example}. Each task \(k\) might involve navigating according to a different set of start and end poses while still avoiding the same static environmental obstacles $C^\star$, which, here, refers to the location of the tree. For each task \(k\), we assume access to expert demonstrations, i.e., trajectories \(\xi = \{(s, a)\}\) that are sampled from an expert policy \(\pi_k^\text{E} \in \Pi\). All such trajectories are assumed to maximize reward \(r_k\) while satisfying the constraint $c^\star(s, a) < \infty $ (or $c^\star(s) < \infty)$.\\ 


\section{Background on Inverse Constraint Learning and Safe Control}
\subsection{Prior Work on Inverse Constraint Learning}
\label{iclpriorwork}
One can think of inverse constraint learning (ICL) as analogous to inverse reinforcement learning (IRL). In IRL, one attempts to learn a reward function that explains the expert agent’s behavior \cite{ziebart2008maximum, ziebart2008navigate, ho2016generative, swamy2021moments, swamy2022sequence, swamy2023inverse, sapora2024evil, ren2024hybrid, wu2024diffusing}. Similarly, in ICL, one attempts to learn the constraints that an expert agent implicitly satisfies. 
The main differentiating factors between prior ICL works come from how the problem is formulated (e.g., tabular vs. continuous states), assumptions on the dynamical system (e.g., stochastic or deterministic), and solution algorithms \citep{liu2024comprehensive}. \citet{liu2024comprehensive} also note that a wide variety of ICL algorithms can be viewed as solving the underlying game multi-task ICL game (MT-ICL) formalized by \citet{NEURIPS2023_124dde49}, which we therefore adopt in for our theoretical analysis. \citet{NEURIPS2023_124dde49}'s formulation of ICL readily scales to modern deep learning architectures with provable policy performance and safety guarantees, broadening the practical relevance of our theoretical findings. We note that our primary focus is not the development of a new algorithm to solve the ICL problem, but on what these methods actually recover.

We now briefly discuss a few notable other prior ICL works. \citet{chou2020learning} formulate ICL as an inverse feasibility problem where the state space is discretized and a safe/unsafe label is assigned to each cell in attempt to recover a constraint that is uniformly correct (which can be impractical for settings with high-dimensional state spaces). 
\citet{scobee2019maximum} adapt the Maximum Entropy IRL (MaxEnt) framework by selecting the constraints which maximize the likelihood of expert demonstrations. This approach was later extended to stochastic models by \citet{mcpherson2021maximum} and to continuous dynamics by \citet{stocking2022maximum}. 
\citet{lindner2024learning} define a constraint set through convex combinations of feature expectations from safe demonstrations, each originating from different tasks. This set is utilized to compute a safe policy for a new task by enforcing the policy to lie in the convex hull of the demonstrations. \citet{hugessen2024simplifying}  note that, for certain classes of constraint functions, single-task ICL simplifies to IRL, enabling simpler implementation.

\subsection{A Game-Theoretic Formulation of Multi-Task Inverse Constraint Learning}
\citet{NEURIPS2023_124dde49}'s MT-ICL formulates the constraint inference problem as a zero-sum game between a policy player and a constraint player and is based on the observation that we want to recover constraints that forbid highly rewarding behavior that the expert could have taken but chose not to. 
Equivalently, the technique can be viewed as solving the following bilevel optimization objective \citep{liu2024comprehensive, qadri2024learning, qadri2023learning, huang2023went}, where, given a current estimate of the constraint at iteration $n$, we train a new constraint-satisfying learner policy for each task $k$. Given these policies, a new constraint is inferred (outer objective) by picking the constraint $\hat{c} \in \mathcal{C}$ that maximally penalizes the set of learner policies relative to the set of expert policies, on average over tasks. This process is then repeated at iteration $n+1$. We refer interested readers to \citet{NEURIPS2023_124dde49} for the precise conditions under which convergence rates can be proved. More formally, let $n$ be the current number of outer iterations performed and let $\pi_k^E$ and $\pi_{k,n}^\star$ denote, respectively, the expert and current learner policy for task $k$. Then, we have:
\begin{align}
    & \text{Outer Objective: }  \hat{c} = \argmax_{c \in \mathcal{C}} \frac{1}{K} \mathbb{E}_{i \sim [n]} \left [ \sum_{k=1}^K J(\pi_{k,i}^\star, c) - J(\pi^\text{E}_k, c) \right ]& \label{bilevel}\\
    & \text{Inner Objective: }    
    \pi_{k,n}^\star = \argmax_{\pi_{k} \in \Pi} J(\pi_k, r_k)  & \nonumber \\
    & \quad \quad \quad \quad  \quad \quad \quad \quad 
  \quad \;  \text{s.t. } J(\pi_k, \hat{c}) \leq \delta \quad \forall \, k \in [K] , \nonumber
\end{align}

where $
J(\pi,f)=
\mathbb{E}_{\xi\sim\pi}\!
\Bigl[\,
   \int_{0}^{T} f\bigl(s(t),a(t)\bigr)\,dt
\Bigr] $ is the value of policy \(\pi\) under a
reward/cost function \(f\in\{r_k,c\}\).  \footnote{We note that in some formulations of ICL (e.g. those in \citet{NEURIPS2023_124dde49}), the RHS of the inner constraint is $J(\pi_E, \hat{c})$. We use a fixed $\delta$ for simplicity of implementation, but note that our proofs hold for either choice of upper bound.} Here
\(\xi=\{(s(t),a(t))\}_{t\in[0,T]}\) denotes a trajectory sampled from
\(\pi\), with \((s(t),a(t))\) the state–action pair at time \(t\).
We assume each task reward \(r_k\) yields a finite integral and that
the inner optimization in Eq.~\ref{bilevel} is feasible: For
every outer iteration there exists at least one policy that satisfies
the constraint. The inner loop can be solved using a standard constrained RL algorithm, while the outer loop can be solved via training a classifier to maximally discriminate between the state-action pairs visited by the learner policies computed in the inner loop versus the states-action pairs in the demonstrations. 

\subsection{A Brief Overview of Safety-Critical Control}


Safety-critical control (SCC) provides us with a mathematic framework for reasoning about inevitable failure (i.e. constraint violation) in sequential problems. Most critically for our purposes, SCC differentiates between a \textit{failure set} (the set of states for which failure has already happened) and a \textit{backward reachable tube (BRT)} (the set of states from which failure is inevitable as we have made a mistake we cannot recover from). Connecting back to ICL, observe that the safe expert demonstrations can never pass through their BRT, as it is impossible to avoid violating the true constraint under their own dynamics. 
Formally understanding BRTs will help us precisely understand why the constraint we infer with ICL does not generally equal the true constraint $c^\star$. In particular, we will show in Section~\ref{problemstatement} that in the best-case, $\hat{c}$ approximates the BRT rather than the true failure set. We now provide an overview of BRTs.

\noindent \textbf{Backward Reachable Tube (BRT).}
%
In safe control, the set defined by the true constraint $c^*$ and denoted by $C^\star = \{ s \in \mathcal{S} \; \vert \; 1[c^\star(s)=\infty] \}$ is generally referred to as the \textit{failure set} and is often denoted by $\mathcal{F}$ in the literature. If we know the failure set \textit{a priori}, $\mathcal{F} \subset \mathcal{S}$, we can  characterize and solve for the \textit{safe set}, $S^{\text{safe}} \subseteq \mathcal{S}$: a subset of states from which if the robot starts, there exists a control action $a$ it can take that guarantees it can avoid states in $\mathcal{F}$ despite a worst-case disturbance $d$. 
Let the maximal safe set and the corresponding minimal unsafe set be: 
\begin{align}
    & S^{\text{safe}} := \{ s_0 \in \mathcal{S} \; | \; \exists \pi_a; \forall \pi_d \; | \; \forall t \geq 0, \xi^{\pi_a, \pi_d}_{s_0}(t) \notin \mathcal{F} \} \\ 
    & S^{\text{unsafe}} := (S^{\text{safe}})^c = \text{BRT}(\mathcal{F}) 
    \label{eq:safe-and-unsafe-sets}
\end{align}
where $\mathcal{S}$ is the state space, $\pi_a$ and $\pi_d$ are respectively the control and disturbance policies, $\xi_{s_0}^{\pi_a, \pi_d}$ is the system trajectory starting from state $s_0$ and following $\pi_a, \pi_d$, and ``$(\cdot)^{c}$" indicates that the set complement of $S^{\text{safe}}$ is the \textit{unsafe set} $S^{\text{unsafe}} \subseteq \mathcal{S}$. 
In the safe control community, the unsafe set is often called the \textit{Backward Reachable Tube} (BRT) of the failure set (i.e., the true constraint) $\mathcal{F}$ \citep{mitchell2005time, bansal2017hamilton}. 
In general, obtaining the BRT is computationally challenging but has been studied extensively by the Hamilton-Jacobi (HJ) reachability \citep{mitchell2005time, margellos2011hamilton} and control barrier functions (CBFs) \citep{ames2019control, xiao2021high} communities.

We ground this work in the language of HJ reachability for a few reasons. First, HJ reachability is guaranteed to return the \textit{minimal} unsafe set -- when studying the best constraint that ICL could ever recover, the BRT obtained via HJ reachability gives us the tightest reference point.  
Second, HJ reachability is connected to a suite of numerical tools for computationally constructing the unsafe set given the true failure set and is compatible with arbitrary nonlinear systems, nonconvex failure sets $\mathcal{F}$, and also incorporate robustness to exogenous disturbances. 

\noindent \textbf{Hamilton-Jacobi (HJ) Reachability} 
computes the unsafe set from Eq.~\ref{eq:safe-and-unsafe-sets}
by posing a robust optimal control problem. Specifically, we want to determine the closest our dynamical system $\dot{s} = f(s,a,d, t)$
could get to $\mathcal{F}$ over some time horizon $t \in [-T,0]$ (where $T$ can approach $\infty$) assuming the control expert tries their hardest to \textit{avoid} the constraint and the disturbance tries to \textit{reach} the constraint.  
This can be expressed as a zero-sum differential game between the control $a$ and disturbance $d$, in which the control tries to steer the system away from failure region while the disturbance attempts to push it towards the unsafe states. Solving this game is equivalent to solving the Hamilton-Jacobi-Isaacs Variational Inequality (HJI-VI) \citep{margellos2011hamilton, fisac2015reach}:
\begin{align}
& \min\Big\{h(s) - V(s,t), ~\nabla_t V(s,t) + \max_{a \in \mathcal{A}} \min_{d \in D} \nabla_{s} V(s,t) \cdot f(s, a, d, t) \Big\} = 0 \label{eq:HJIVI}  \\ 
& V(s, 0) = h(s), \quad t \leq 0 \nonumber
\end{align}
where $h(s)$ is a Lipschitz function encoding the failure set $\mathcal{F} =\{ s \; | \; h(s) \leq 0 \}$, and $\nabla_t V(s,t), \nabla_s V(s,t)$ are respectively, the gradients with respect to time and state. 
The HJI-VI in \eqref{eq:HJIVI} can be solved via dynamic programming and high-fidelity grid-based PDE solvers \citep{mitchell2004toolbox} or function approximation \citep{fisac2019bridging, bansal2021deepreach, hsu2023isaacs}. 
As $t \rightarrow -\infty$, the value function no longer changes in time and we obtain $V^\star(s)$ which represents the infinite time control-invariant BRT, which can be extracted via the sub-zero level set of the value function:
\begin{equation}
    \text{BRT}(\mathcal{F}) := S^{\text{unsafe}} = \{s \in \mathcal{S} : V^\star(s) < 0\}.
\end{equation}

\section{What Are We Learning in ICL?}
One might naturally assume that an ICL algorithm would recover the true constraint  $c^{\star}$ (e.g. the exact location of the tree, illustrated in Fig. \ref{fig:simple-example}b) that the expert optimizes under.
\label{problemstatement}
However, we now prove that the set $\hat{C}$, induced by the inferred constraint $\hat{c}$, is equivalent to the BRT of the failure set, $\text{BRT}(\mathcal{F})$, where $\mathcal{F} \equiv C^\star$. 
In other words, we prove that constraint inference ultimately learns a dynamics-conditioned \textit{unsafe set} instead of the dynamics-independent true constraint. 

Throughout this section, we assume we are in the single-task setting ($K=1$) for simplicity and drop the associated subscript. Let $P(\cdot): \Pi \to \mathbb{R}$ be a function which maps a policy $\pi$ to some performance measure. For example, in our preceding formulation of multi-task ICL, we had set $P_k(\pi_k) = J(\pi_k, r_k)$. We begin by proving that relaxing the failure set to its BRT does not change the set of solutions to a safe control problem. This implies that, from safe expert demonstrations alone, we cannot differentiate between the true failure set and its BRT.


\begin{lemma}
Consider an expert who attempts to avoid the ground-truth failure set $\mathcal{F}$ under dynamics $\dot{s} = f(s, a, d, t)$ while maximizing performance objective $P: \Pi \rightarrow \mathbb{R}$:
\begin{align}
    & \pi^\star_a = \argmax_{\pi \in \Pi} P(\pi) \label{original-prob} \\
    & \text{s.t. } J(\pi, \mathbbm{1}[\cdot \in \mathcal{F}]) = 0 . \nonumber
\end{align}
Also consider the relaxed problem below, where the expert avoids the $\text{BRT}$ of the failure set $\mathcal{F}$:
\begin{align}
    & \pi^\star_b = \argmax_{\pi \in \Pi} P(\pi) \label{relaxed} \\
    & \text{s.t. } J(\pi, \mathbbm{1}[\cdot \in \text{BRT}(\mathcal{F})]) = 0.
    \nonumber
\end{align} 
Where $\mathbbm{1}[\cdot \in \mathcal{F}]$ and $\mathbbm{1}[\cdot \in \text{BRT}(\mathcal{F})]$ are indicator functions that assign the value 1 to states $s \in \mathcal{F}$ and $s \in \mathcal{\text{BRT}(F)}$ respectively and the value 0 otherwise.   Then, the two problems \ref{original-prob} and \ref{relaxed} have equivalent sets of solutions, i.e. 
\begin{align}
    \pi_a^\star = \pi_b^\star.
\end{align}
\end{lemma} 

\begin{pf} By the definition of the BRT in Eq.~\ref{eq:safe-and-unsafe-sets}, we know that $\forall s \in \text{BRT}(\mathcal{F})$, any trajectory $\xi^{\pi(\cdot)}_s(t)$ that starts from state $s$ and then follows any policy $\pi$ with $\pi \in \pi^\star_a$ is bound to enter the failure set. Thus, we know that no policy in $\pi^\star_a$ will generate trajectories that enter the BRT, i.e. $\forall \pi \in \pi^{\star}_a$, $J(\pi, \mathbbm{1}[\cdot \in \text{BRT}(\mathcal{F})]) = 0$. This implies that $\pi^{\star}_a \subseteq \pi^{\star}_b$. Next, we observe that $\mathcal{F} \subseteq \text{BRT}(\mathcal{F})$. This directly implies that $\forall \pi \in \pi^{\star}_b$, $J(\pi, \mathbbm{1}[\cdot \in \mathcal{F}]) = 0$, which further implies that $\pi^{\star}_b \subseteq \pi^{\star}_a$. Taken together, the preceding two claims imply that $\pi^{\star}_a = \pi^{\star}_b$.\end{pf}

Building on the above result, we now prove an equivalence between solving the ICL game and BRT computation. First, we define $P_{\mathbb{H}}$ as the entropy-regularized cumulative reward, i.e.
\begin{align}
    P_{\mathbb{H}}(\pi) \triangleq J(\pi, r) + \mathbb{H}(\pi), 
\end{align}
where  $\mathbb{H}(\pi) = \mathbb{E}_{\xi \sim \pi}\left[\int_t^T -\log\pi(a_t | s_t) dt \right] $ is causal entropy \citep{massey1990causality, ziebart2010modeling, kramer1998directed}. We now prove that a single iteration of \textit{exact, entropy-regularized} ICL recovers the BRT.

\begin{theorem}
\label{maintheorem}
Define $\pi^\text{E} = \argmax_{\pi \in \Pi} P_{\mathbb{H}}(\pi)$ s.t. $J(\pi, c^{\star}) \leq 0$ as the (unique, soft-optimal) expert policy. Let $\hat{c}_0 = 0, \forall s \in \mathcal{S}$, and define $\hat{\pi}_0 = \argmax_{\pi \in \Pi} P_{\mathbb{H}}(\pi)$ s.t. $J(\pi, \hat{c}_0) \leq 0$ as the (unique) soft-optimal solution to the first inner ICL problem. Next, define 
 \begin{align}
    \hat{c}_1  = \argmax_{c \in \{\mathcal{S} \to \mathbb{R}\}} \mathbb{E}_{s^+ \sim \hat{\pi}_0, s^-\sim \pi^\text{E}} [\log(\sigma(c(s^+) - c(s^-)))],
\end{align}
where $\sigma(x) = \frac{1}{1 + \exp(-x)}$,
as the optimal classifier between learner and expert states. Then,
\begin{equation}
    \hat{C} = \{ s \in \mathcal{S} \; \vert \; \mathbbm{1}[\hat{c}_1(s)=\infty] \} = \text{BRT}(\mathcal{F}).
\end{equation}

\end{theorem}

\begin{pf}
We use $\rho_{\pi}$ to denote the visitation distribution of policy $\pi$: $ \rho^{\pi}(s') = \mathbb{E}_{s \sim \pi}[1[s = s']]$. First, we observe that under a $c_0$ that marks all states as safe, the inner optimization reduces to a standard, unconstrained RL problem. It is well known that the optimal classifier for logistic regression is
\begin{equation}
    \hat{c}_1(s) = \log \left(\frac{\rho^{\hat{\pi}_0}(s)}{\rho^{\pi^\text{E}}(s)}\right).
\end{equation}

We then recall that because of the entropy regularization, $\pi^{\star}_0$ has support over all trajectories that aren't explicitly forbidden by a constraint \citep{phillips2008modeling, ziebart2008maximum}. Because there is no constraint at iteration $0$, this implies that $\forall s \in \mathcal{S}$, $\rho^{\hat{\pi}_0}(s) > 0$.

By construction, we know $\pi^\text{E}$ will never enter the failure set $\mathcal{F}$. By our preceding lemma, we know it will also never enter the BRT. This implies that $\forall s \in \text{BRT}(\mathcal{F}),$ $\rho^{\pi^\text{E}}(s) = 0$. Given these are the only moment constraints we have to satisfy, this also implies that $\pi^\text{E}$ will have full support over all states that aren't in $\text{BRT}(\mathcal{F})$, i.e. $\forall s \in \mathcal{S} \backslash \text{BRT}(\mathcal{F})$, $\rho^{\pi^\text{E}}(s) > 0$.

Taken together, this means that $\forall s \in \text{BRT}(\mathcal{F}),$ $\hat{c}_1(s) = \infty$; and $\forall s \in \mathcal{S} \backslash \text{BRT}(\mathcal{F}),$ $\hat{c}_1(s) < \infty$. \\ Thus, $ \{ s \in \mathcal{S} \; \vert \; \mathbbm{1}[\hat{c}_1(s)=\infty] \} \equiv \text{BRT}(\mathcal{F})$. %
\end{pf}

In summary, assuming access to a perfect solver, the ICL procedure recovers the BRT of the failure set, rather than the failure set itself under fairly mild other assumptions. Before we discuss the implications of this observation, we experimentally validate how well ICL recovers the BRT.

\textbf{Remark (Multi-task Case).}
Theorem \ref{maintheorem} relies on the entropy term to give the expert policy
full support over the safe region which is difficult to guarantee in practical scenarios.  With $K$ tasks, this requirement is relaxed because the
combined demonstrations cover the state space more thoroughly.  The
result itself carries over unchanged: if all tasks share the same state
space, dynamics, and implicit constraint, we can replace the single expert
density in the proof by the aggregate
\begin{align}
\rho^{\text E}(s)=\sum_{k=1}^{K}\rho^{\pi^{\text E}_{k}}(s).
\end{align}
Any state inside the backward-reachable tube has $\rho^{\text E}(s) = 0$, so the classifier again labels it with
$\hat{c}_1(s)=\infty$, and the inferred constraint coincides with the BRT.



\section{Experimental Validation of BRT Recovery}

Our theoretical statements assumed access to a perfect ICL solver. We now empirically demonstrate that even when this assumption is relaxed, we see that $\hat{c}$ approximates the BRT.

\subsection{Dynamical System}

In our experiments, we select a low-dimensional but dynamically-nontrivial system that enables us to effectively validate our theoretical analysis through empirical observation. 

Specifically, we investigate a Dubins' car-like system with a state defined by position and heading: $s = (x, y, \theta)$. The continuous-time dynamics are modeled as:
\begin{align}
f(s,a,d,t) = f_0(s,t) + G_u(s,t) \cdot a + G_d(s,t) \cdot d.     
\end{align}

The robot's dynamics are influenced by its control inputs which are linear and angular velocity $a := [v, \omega] \in \mathcal{A}$, an extrinsic disturbance vector $d := [d^x, d^y] \in \mathcal{D}$ acting on $x$ and $y$, and open loop dynamics $f_0 = \begin{bmatrix}
    v_{\text{nominal}} \cos(\theta),  
    v_{\text{nominal}} \sin(\theta) , 
    0
\end{bmatrix}^T$ with nominal speed $v_\text{nominal}=0.6$. 
Finally, 
$$G_u = \begin{bmatrix}
    \cos(\theta) & 0 \\ 
    \sin(\theta) & 0 \\
    0 & 1
\end{bmatrix}, G_d = \begin{bmatrix}
    1 & 0 \\ 
    0 & 1 \\
    0 & 0
\end{bmatrix}$$ 
are respectively the control and disturbance Jacobians. 

In our experiments, we study two dynamical systems: Model 1, an \textbf{agile} system with strong control authority $v \in [-1.5, 1.5]$ and $\omega \in [-1.5, 1.5]$, and Model 2, a \textbf{non-agile} system with less control authority, $v \in [-0.7, 0.7]$ and $\omega \in [-0.7, 0.7]$. 
In all experiments, $d^i \in [-0.6, 0.6], i \in {x,y}$.
This setup was selected to demonstrate how constraint inference can effectively ``hide'' the BRT when the dynamical system is sufficiently agile (see subsection \ref{62}).

 \subsection{Constraint Inference Setup}

 We use the MT-ICL algorithm developed by \citet{NEURIPS2023_124dde49}. In our setup, each task $k$ involves navigating a robot from a specific start state $s_k$ to a goal state $g_k$ without colliding with a circular obstacle of radius 1, centered at the origin. This obstacle defines the ground-truth constraint in the expert demonstrator’s mind, $c^\star$. We model the constraint to be a function of only the state  $\hat{c}: s \rightarrow [-\infty, \infty]$. Note that in practice, we constraint the output of $\hat{c}$ to be in the range $[-1, 1]$. Let $\mathcal{C}$ denote the function class of 3-layer MLPs, and $\Pi$ denote the set of actor-critic policies where both actor and critic are 2-layer MLPs. All networks use a hidden size of 128 with ReLU activations. The inner constrained RL loop is solved using a penalty-based method where a large negative reward is applied upon violation of $\hat{c}$. For each task, we train an expert policy using PPO \citep{schulman2017proximal}, implemented via the Tianshou library \citep{tianshou}, with perfect knowledge of the environment (i.e., obstacle location). PPO includes entropy regularization as assumed in Section~\ref{problemstatement}. We collect approximately 200 expert demonstrations across diverse start and goal configurations to form two training sets, $\mathcal{D}_\text{agile}$ and $\mathcal{D}_\text{non-agile}$. Each dataset is then used to train MT-ICL (Eq.~\ref{bilevel}) for 5 epochs, using only access to the demonstrations. All models were trained on a single NVIDIA RTX 4090 GPU. Further implementation details are available in the codebase.

 \subsection{BRT Computation}

We solve for the infinite-time avoid BRT using an off-the-shelf solver of the  HJI-VI PDE (Eq. \ref{eq:HJIVI}) implemented in JAX \citep{hj_reachability}. We encode the true circular constraint via the signed distance function to the obstacle: $h(s) := \{s : || 
\begin{bmatrix}
s^x \\ s^y    
\end{bmatrix} - 
\begin{bmatrix}
o^x \\ o^y    
\end{bmatrix}||^2_2 < r^2\}$. 
We initialize our value function with this signed distance function $V(0, s) = h(s)$ and discretize full the state space $(x,y,\theta)$ into a grid of size $200\times200\times200$. We run the solver until convergence. 

\subsection{Results.}
We now discuss several sets of experimental results that echo our preceding theory.

\begin{figure}[t!]
    \centering    \includegraphics[width=0.92\linewidth]{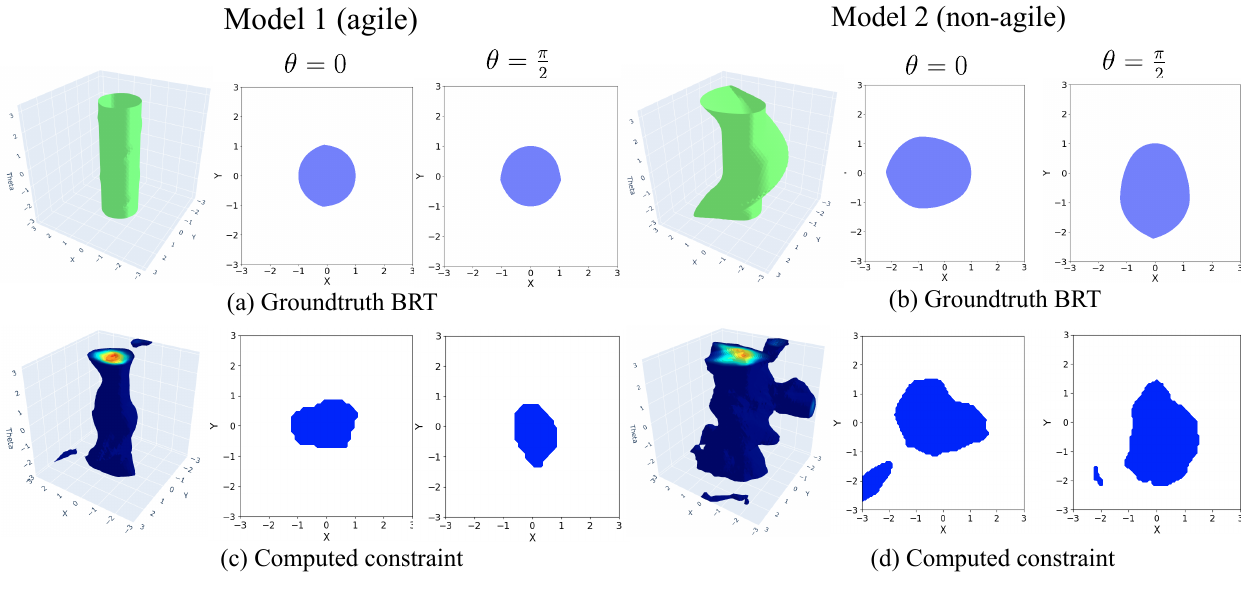}
    \caption{(a) and (b) show the Backward Reachable Tube (BRT) while (c) and (d) show the inferred constraint via ICL for both the agile (model 1) and non-agile (model 2) systems.}
    \label{fig:brt}
\end{figure}

\subsubsection{ICL Recovers an Approximation of the BRT}

\begin{wrapfigure}{r}{0.5\textwidth}
  \vspace{0em}
  \begin{center}
    \includegraphics[width=0.5\textwidth]{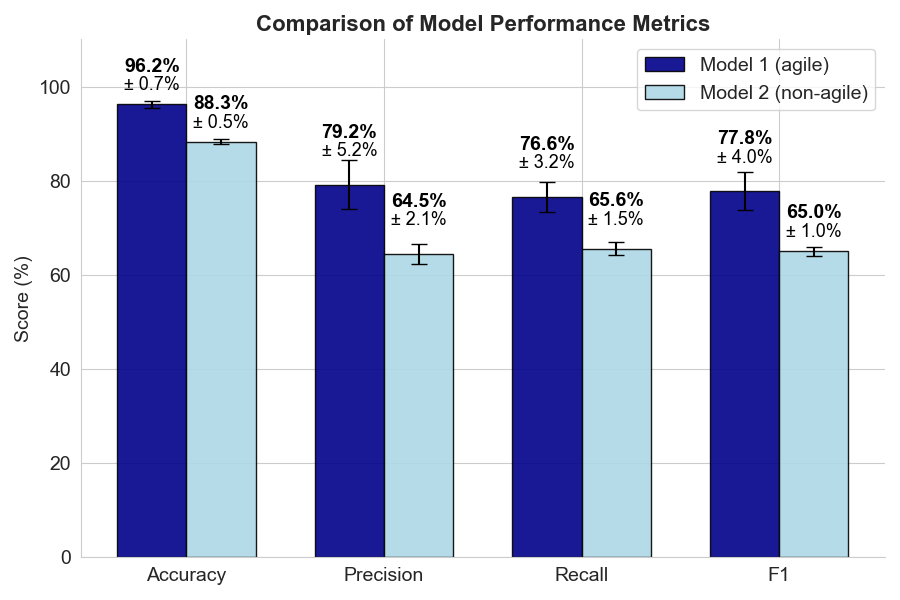}
  \end{center}
  \vspace{0em}
  \caption{Classification metrics (mean and standard deviation averaged over three different seeds) for the estimated unsafe set \(\hat{C}\) vs. true failure set $C^\star$. The plot presents performance scores (Accuracy, Precision, Recall, and F1) for the two models, with error bars indicating the variability across the three seeds.}
  \label{fig:classification_metrics}
  \vspace{-5em}
\end{wrapfigure}

First, we compute the ground truth BRTs for each model by solving the HJB PDE in Eq. \ref{eq:HJIVI}. Figures~\ref{fig:brt}a and \ref{fig:brt}b show how each model induces a different BRT, with the BRT growing larger as the control authority decreases. This indicates that less agile systems result in  a larger set of states that are bound to violate the constraint. We then use MT-ICL to compute \(\hat{c}_{\text{agile}}(s)\) and \(\hat{c}_{\text{non-agile}}(s) \), the inferred constraint for the agile and non-agile systems respectively.
In figures~\ref{fig:brt}c and \ref{fig:brt}d, we visualize the constraints by computing the level sets \(\hat{c}_{\text{agile}}(s) > 0.6\) and \(\hat{c}_{\text{non-agile}}(s) > 0.6\), indicating a high probability of a state \(s\) being unsafe. We observe an empirical similarity between the ground truth BRTs and the learned constraints. Additionally, we report quantitative metrics for our classifiers in Fig. \ref{fig:classification_metrics}, averaged over three different seeds. These quantitative and qualitative results support our argument that the inferred constraint $\hat{c}_{\text{agile}}(s)$ and $\hat{c}_{\text{non-agile}}(s)$  are indeed approximations of the BRTs for model 1 (agile dynamics) and model 2 (non-agile dynamics) respectively. We note that the classification errors can be attributed to limited expert coverage in certain parts of the state space. This limitation arises from capping the number of start-goal states at $K\approx200$ (i.e., the total number of tasks) due to the high computational cost of the inner MT-ICL loop, which involves training a full RL model for each task.

\subsubsection{ICL Can ``Hide'' the BRT When the System is Agile}
\label{62}
Agile systems are commonly used in the existing ICL literature, leading to the impression that the set inferred from constraint $\hat{c}$ (the set $\hat{C} = \{s \in \mathcal{S} \; | \; 1[\hat{c}(s) =\infty \}$) is always equal to the failure set $\mathcal{F} = C^\star$. However, we note that this equivalence holds only when $\text{BRT}(\mathcal{F}) \approx \mathcal{F}$—i.e., when the system possesses sufficient control authority to ``instantaneously quickly'' steer away from the failure set or ``instantaneously'' stop before entering failure (e.g. model 1 in Fig. \ref{fig:brt}). For general dynamics (e.g. model 2 in Fig. \ref{fig:brt}), $\hat{C} \neq \mathcal{F}$ when $\text{BRT}(\mathcal{F}) \neq \mathcal{F}$.

\subsubsection{The Constraint Inferred via ICL Doesn't Necessarily Generalize Across Dynamics}
The fact that ICL approximates a backwards reachable tube has direct implications on the transferability of the learned constraint across different dynamics: since the BRT is inherently conditioned on the dynamics, the constraint computed by ICL will be as well. We discuss the implications of this observation on downstream policy optimization that uses the inferred constraint from ICL. 

Specifically, we study the following general formulation for learning a policy for dynamical system model $a$, using an ICL-derived constraint derived from a \textit{different} dynamical system model, $b$: 
\begin{align}
    & \pi^\star_{a \mid \text{BRT}_b} = \argmax_{\pi \in \Pi} P(\pi) \label{general_formulation_1} \\
    & \text{s.t. } J(\pi, \mathbbm{1}[\cdot \in \text{BRT}_b]) = 0. \nonumber
\end{align}

We compare this solution against a policy learned for model $a$ using a constraint derived from demonstrations given on the \textit{same} dynamical system model, $a$. We assume that for the given system (model a) and constraint set ($\text{BRT}_b$), there exists a feasible policy that can satisfy the constraint. If this is not the case, the optimization problem is infeasible, highlighting another important limitation of naive constraint transfer across dynamics.

Let $\mathbbm{1}[\cdot \in \text{BRT}_a]$, $\mathbbm{1}[\cdot \in \text{BRT}_b]$ be indicator functions representing state membership in the respective BRTs.
For this analysis, let dynamical system models $a$ and $b$ share the same state space, e.g.,  $\mathcal{S}= \{(x, y, \theta) \}$, and dynamical system evolution, e.g., a 3D Dubin's car model where the robot controls both linear and angular velocity. However, they will differ in their control authority, i.e., the action space $\mathcal{A}$. Let $a, b \in \{M_<, M, M_>\}$ be the possible models we could analyze: 

\begin{itemize}
\item $M_<$ denote a non-agile system; for example $\mathcal{A}$ significantly limits how fast the system can turn. 
\item $M$ is a moderately agile system. 
\item $M_>$ is an agile system with sufficient control authority to always avoid the failure set; for example, $\mathcal{A}$ can turn extremely fast and stop instantaneously. 
\end{itemize}

\begin{wrapfigure}{r}{0.4\textwidth}
  \vspace{-1.8em}
  \begin{center}
    \includegraphics[width=0.4\textwidth]{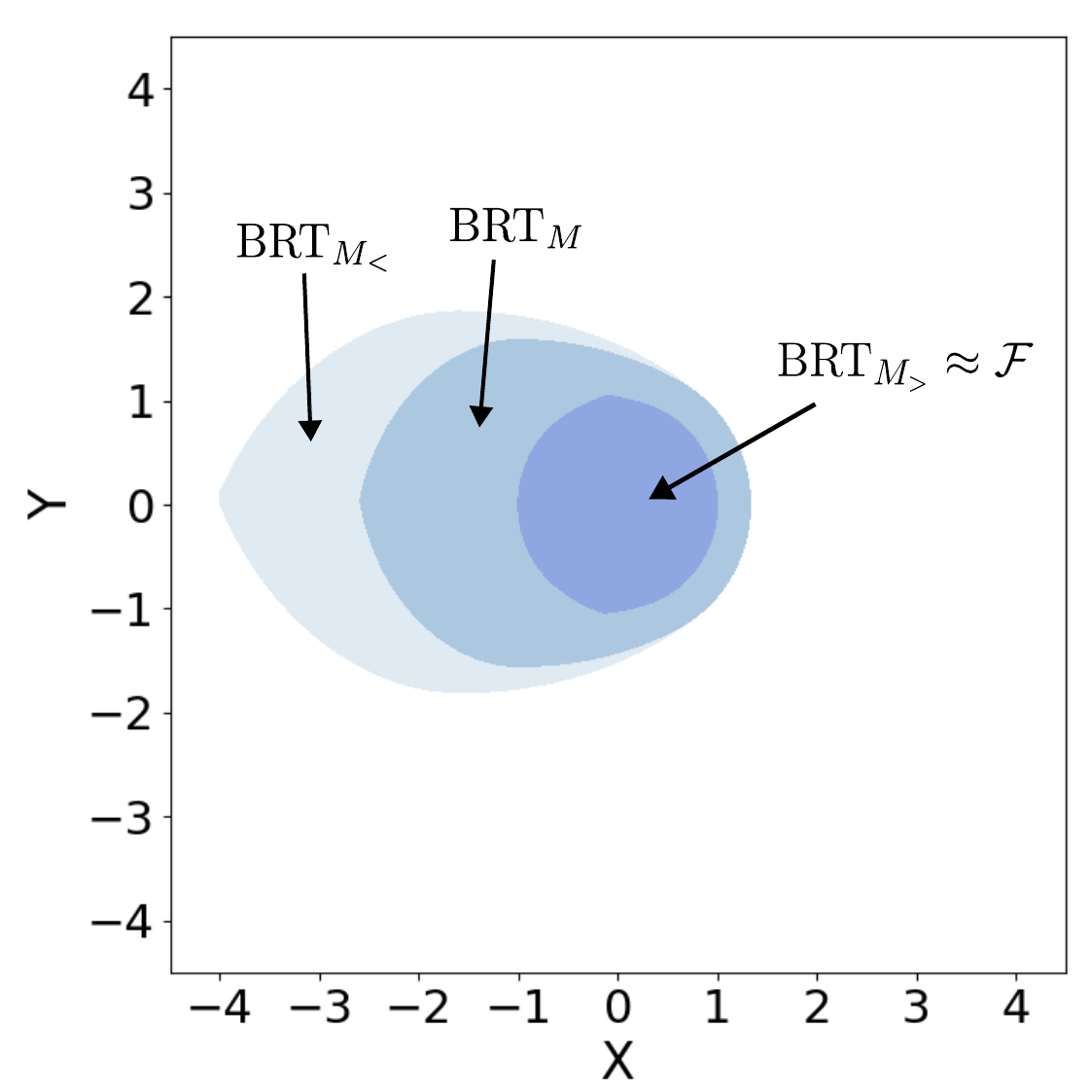}
  \end{center}
 \vspace{-1.2em}
  \caption{Illustration of the relationship between the three BRTs we want to analyze: $\text{BRT}_{M_>}$, $\text{BRT}_{M}$ and $\text{BRT}_{M_<}$. They satisfy the relationship  in Eq. \ref{brtanalysisrel}. }
  \label{fig:BRTanalysis}
 \vspace{-1.5em}
\end{wrapfigure}
The corresponding unsafe sets for each of these system models satisfy the following relation (see visualization in Fig. \ref{fig:BRTanalysis}):
\begin{align}
\mathcal{F} \equiv \text{BRT}_{M_>} \subset \text{BRT}_{M} \subset \text{BRT}_{M_<}
\label{brtanalysisrel}
\end{align}
\begin{align}
    & \pi^\star_{a \mid \text{BRT}_a} = \argmax_{\pi \in \Pi} P(\pi) \label{general_formulation_2} \\
    & \text{s.t. } J(\pi, \mathbbm{1}[\cdot \in \text{BRT}_a]) = 0. \nonumber
\end{align}

Finally, we define the operator $g_{\text a}:\mathcal{P}(\mathcal{S}) \rightarrow \mathcal{P}(\mathcal{S})$, where $\mathcal{P}(\mathcal{S})$ is the power set of $\mathcal{S}$. Here, $g_a$ takes as input \textit{any} set of states that must be avoided and outputs the corresponding BRT for this failure set under dynamical system model $a$. For example, $g_{M_<}(\text{BRT}_{M_>})$ is the BRT computed for model $M_<$ with $\text{BRT}_{M_>}$ as the target initial set (i.e. $V(s, 0)$ in Eq. \ref{eq:HJIVI} is defined such that $V(s, 0) < 0$ when $s \in \text{BRT}_{M_>}$) .

\textbf{Transferring the Learned Constraint from Agile to Less-Agile Systems.}
This scenario is equivalent to setting  model $a=M$ or $a=M_<$ and $b=M_>$ in Eq. \ref{general_formulation_1}. 
Since the constraint learned for model $M_>$ is equivalent to the failure set (i.e. $\text{BRT}_{M_>} \equiv \mathcal{F}$), then by Lemma 4.1, the policy which satisfies the inferred constraint $\pi^\star_{a \mid \text{BRT}_b}$ will not be over-conservative. In other words, $\pi^\star_{a \mid \text{BRT}_b}$ will be approximately the same as the policy obtained under the BRT computed on the \textit{same} dynamical system, $\pi^\star_{a \mid \text{BRT}_a}$. 

\textbf{Transferring the Learned Constraint from Non-Agile to more Agile Systems.}
This scenario is equivalent to setting  model $a=M$ or $a=M_>$ and $b=M_<$, or setting $a=M_>$ and $b=M$ in Eq. \ref{general_formulation_1}. 
Since the inferred constraint $\text{BRT}_b$ was retrieved from a less agile system, we know that it is larger than the failure set ($\mathcal{F}$) and larger than the BRT of the target system $a$ ($\text{BRT}_a$) that we want to do policy optimization with. 
This means that if we use the constraint $\text{BRT}_b$ during policy optimization with a target system that is more agile, rollouts from the resulting policy $\pi^\star_{a \mid \text{BRT}_b}$ will have to avoid \textit{more} states than the failure set $\mathcal{F}$ or the target system's true unsafe set, $\text{BRT}_a$. 
Mathematically, rollouts generated from the optimized policy $\pi^\star_{a \mid \text{BRT}_b}$ will implicitly satisfy  $g_{a}(\text{BRT}_{b})$, which is a superset of $\text{BRT}_{a}$, and hence, yields an overly conservative solution compared to Eq. \ref{general_formulation_2}.  

\textbf{Transferring the Learned Constraint from a Moderately-Agile to a Non-Agile System.} This scenario is equivalent to setting  model $a=M_<$ and model $b=M$ in Eq. \ref{general_formulation_1}. In this case, rollouts generated from the optimized policy $\pi^\star_{a \mid \text{BRT}_b}$ will implicitly satisfy  $g_{a}(\text{BRT}_{b})$ which is a superset of $\text{BRT}_{a}$. Again, this means that the robot will avoid states from which it could actually remain safe leading to suboptimal policies compared to rollouts of the solution policy to Eq. \ref{general_formulation_2}.

\section{Conclusion, Implications, and Future Work}
\vspace{-1em}

\label{sec:conclusion}

In this work, we have identified that inverse constrained learning (ICL), in fact, approximates the backward reachable tube (BRT) using expert demonstrations, rather than the true failure set. We now argue that this observation has a positive impact from a \textit{computational} perspective and a negative impact from a \textit{transferability} perspective.

\textbf{Implications.} First, we note that we can add ICL algorithms to the set of computational tools available to us to calculate BRTs, given a dataset of safe demonstrations, without requiring prior knowledge of the true failure set. Computing a BRT is the first step in many downstream safe control synthesis procedures of popular interest. We also note that having access to a BRT approximator can help speed up policy search, as the set of policies that do not violate the constraint is a subset of the full policy space. Thus, a statistical method should take fewer samples to learn the (safe) optimal policy with this knowledge. However, any BRT (inferred by ICL or otherwise) is dependent on the dynamics of the system and hence cannot be easily used to learn policies on different systems without care. 
In this sense, learning a BRT rather than the failure set is a double-edged sword.

We note that in some sense, learning a BRT rather than a failure set is analogous to learning a value function rather than a reward function. In particular, the BRT is the zero sublevel set of the safety value function. While value functions make it easier to compute an optimal policy, their dynamics-conditionedness makes them more difficult to transfer across problems.

We also note that the above observations are somewhat surprising from the perspective of inverse reinforcement learning, where one of the key arguments for learning a reward function is transferability across problems \citep{ng2000algorithms, swamy2023inverse, sapora2024evil}. However, such transfer arguments often implicitly assume access to a set of higher-level features which are independent of the system's dynamics on top of which rewards are learned, rather than the raw state space as used in the preceding experiments for learning constraints. Thus, another approach to explore is whether the transferability of constraints would increase if we learn constraints on top of a set of features which are \textit{1)} designed to be dynamics-agnostic and \textit{2)} for which the target system is able to match the behavior of the expert system, as is common in IRL practice \citep{ziebart2008maximum}.




\textbf{Future Work.} Regardless, an interesting direction for future research involves recovering the true constraint  (i.e., the failure set \(\mathcal{F}\)) using constraints that were learned for different systems with varying dynamics. This process is synonymous to removing the dependence of the constraint on the dynamics by integrating over (i.e., marginalizing) the dynamical variables. This could allow disentangling the dynamics and semantics parts of the constraint, allowing better generalization and faster policy search independent of system dynamics. A potential approach to doing so would be to collect expert demonstrations under a variety of dynamics, learn a constraint for each, and then return an aggregate constraint that is the minimum of the learned constraints, implicitly computing an intersection of the BRTs.  Such an intersection would approximate the true failure set. 

\subsubsection*{Acknowledgments}
\label{sec:ack}
We thank Harley Wiltzer for feedback on our initial draft. MQ and MK were partially supported by the Office of Naval Research (ONR) grant N00014-24-1-2272. GKS is supported by an STTR grant. AB is partially supported by National Science Foundation (NSF) M3X Award \#2246447. 

 .


\newpage
\appendix





\label{sec:ack}



\bibliography{main}
\bibliographystyle{rlj}



\end{document}